# A Fuzzy Topsis Multiple-Attribute Decision Making for Scholarship Selection

**Shofwatul 'Uyun\*[1], Imam Riadi[2]**
[1]Informatics Department, State Islamic University of Sunan Kalijaga
Jl. Marsda Adisucipto No. 1 Yogyakarta 55281, Telp. (0274) 519739, Fax (0274) 540971
[2]Information System Department, University of Ahmad Dahlan
Jl, Prof.Dr.Soepmomo, Janturan, Yogyakarta, Telp (0274) 563515, Fax (0274) 564604
e-mail: shofwatul.uyun@uin-suka.ac.id\*[1], imam_riadi@uad.ac.id[2]

***Abstrak***

*Biaya pendidikan semakin mahal, banyak mahasiswa mengajukan beasiswa. Ratusan bahkan ribuan formulir pengajuan beasiswa harus diseleksi oleh sponsor. Permasalahan tersebut bertujuan untuk memilih beberapa alternatif terbaik berdasarkan beberapa atribut (kriteria) yang digunakan. Dalam rangka pengambilan keputusan pada permasalahan yang bersifat fuzzy dapat digunakan Fuzzy Multiple Attribute Decision Making (FMADM). Pada penelitian ini dilakukan pemodelan menggunakan Unified Modelling Language (UML) pada FMADM dengan metode TOPSIS dan Weighted Product untuk menyeleksi calon penerima beasiswa akademik dan non akademik di Universitas Islam Negeri Sunan Kalijaga. Data yang digunakan adalah data fuzzy dan crisp. Hasil penelitian menunjukkan bahwa Metode TOPSIS dan Weighted Product pada FMADM dapat digunakan untuk seleksi beasiswa. Hasil seleksi merekomendasikan mahasiswa yang memiliki tingkat kelayakan paling tinggi untuk mendapatkan beasiswa berdasarkan nilai preferensi yang dimiliki.*

***Kata kunci****: Fuzzy Multiple Attribute Decision Making, TOPSIS, Weighted Product, Scholarship*

***Abstract***

*As the education fees are becoming more expensive, more students apply for scholarships. Consequently, hundreds and even thousands of applications need to be handled by the sponsor. To solve the problems, some alternatives based on several attributes (criteria) need to be selected. In order to make a decision on such fuzzy problems, Fuzzy Multiple Attribute Decision Making (FMDAM) can be applied. In this study, Unified Modeling Language (UML) in FMADM with TOPSIS and Weighted Product (WP) methods is applied to select the candidates for academic and non-academic scholarships at Universitas Islam Negeri Sunan Kalijaga. Data used were a crisp and fuzzy data. The results show that TOPSIS and Weighted Product FMADM methods can be used to select the most suitable candidates to receive the scholarships since the preference values applied in this method can show applicants with the highest eligibility.*

## 1. Introduction

The national education system defines education as conscious and plans some efforts to create a good learning atmosphere and learning process. Therefore, students can actively develop their potentials so they will have strong religious faith, self-control, strong personality, intellectual, ethics and skills for themselves, society, and country. In line with those purposes are four education visions by UNESCO (United Nation on Education, Scientific and Cultural Organization) in 21$^{st}$ century. Those are (1) learning how to learn, (2) learning how to do, (3) learning how to be, and (4) learning how to live together. In order to support the process, Ministry of Religious Affairs has been offering scholarships for students at UIN Sunan Kalijaga in a regular basis, including scholarships for students with high academic achievements. Some research on application of multi-attribute decision making (MADM) has been widely conducted. In its development, research on MADM is also focus on how the decision makers give their preferences on certain alternatives and criteria [1]. Typically, the decision makers gave numeric weighting preferences to make the computation easier. However, current linguistic preferences are also applied to simplify the decision makers in giving their opinions. For example, the value of alternative A1 is "very good" in the criterion C1, while alternative A1 is "moderate" in criterion C2. The importance level of C1 is "very high", while criterion C2 has a "low" level of importance, and so on. If the preference is given linguistically, then fuzzy logic can be used to help solving the problem. Fuzzy logic is very effective to solve the MADM problem where the given data is ambiguous or presented linguistically [2]. In fact, there are a lot of decisions created in the fuzzy environment [3].

The MADM method is used to solve a case which has several alternatives and priority for various attributes. MADM technique is a popular technique and has widely been used in several fields, including: engineering, economics, managements, transportation planning, etc. Several approaches that have been developed are calculating the weights of MADM problems, such as the eigenvector method and ELECTRE. The paper described the formulation of weighting in MADM case with fuzzy decision matrix was generated by two people [4]. Fuzzy multi attribute decision making (FMADM) has been used to select future lecturer at Department of Computer Science, Faculty of Industrial Technology, Islamic University of Indonesia (UII) using genetic algorithm to find the value of attribute weights. The value is searched through subjective approach. After the weight of every alternative has been found, the grades were processed to determine optimal alternatives; those are the applicants who have been accepted as the future lecturer at Department of Computer Science UII. In addition, the FMADM has also been used to determine the best location for a warehouse (from several alternative locations), using genetic algorithms in finding the value of attribute weights [5], [6].

Fuzzy model is also used to select a project for research and development (R & D) with multi-criteria decision making. The project selection used several qualitative and quantitative criteria. The criteria include cost and some of the obtained advantages if the project was implemented. However, models produced by Ramadan [7] still can not be used in group.In order to anticipate a group assessment, Zhou et al. [8] implemented fuzzy logic in decision support system to assess project produced by students. The project is rated by more than one person with several fuzzy criteria. The best project is a project with the highest membership value. Another method for the decision support system is analytical hierarchy process (AHP) fuzzy. AHP fuzzy can help users to make decisions on both structured and semi structured problems [9]. In addition, [10] fuzzy analytical hierarchy process is also used to help make decisions on the process of multicriteria robot selection. Researchers [11] have described several procedures on a modified technique for order preference by similarity to ideal solution (TOPSIS) method so that the TOPSIS can also be used for a case of decision made in group or multi-criteria group decision making (MCGDM). In this study, TOPSIS algorithm is used in FMADM to asses the eligibility of scholarship recipients and helping the decision maker to make a quick, accurate and objective decision.

TOPSIS algorithm is used to evaluate the results of production processes related to environment. Data used in the algorithm is a crisp data so that the output is a quantitative data. The output data will be evaluated and used as an input for the next process [12]. TOPSIS method is suitable to solve the problem decision making by introducing quantity multiplication operation of triangular fuzzy number. A case study indicates that the method can be applied effectively with less information and the quantitative result is objective and reasonable [13]. Linear programming model for multi attribute group decision making (MAGDM) has been introduced by Xu [14]. The given preference information can be presented in these three distinct uncertain preference structures: interval utility values; interval fuzzy preference relations; and interval multiplicative preference relations. The format of preference information attributes in MAGDM is not uniform. Initially, data gathered from the decision makers with various formats. For that Xu [15] propose a method that can accommodate all of decision makers' proposals. Therefore, a research to develop an unified modeling language (UML) for Fuzzy TOPSIS multiple attribute decision making (FMADM) is needed to assess the eligibility of scholarship recipients and helping the decision maker to make a quick, accurate and objective decision.

## 2. Research Method

Requirement gathering and modeling activities are the steps where the needed materials are collected. Analysis of activity diagram results in some potential actors to become the user of the system under development. In general, the methods which are used for multiple attribute decision making in this study can be shown in Figure 1.

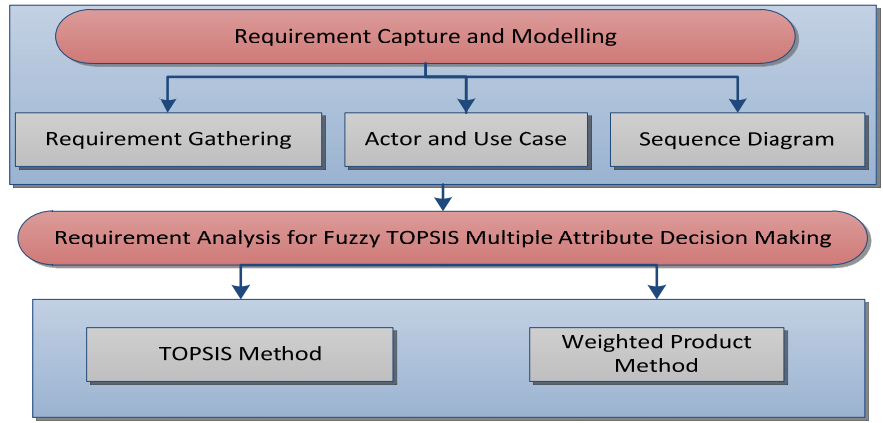

Figure 1. The multiple attribute decision making

The Requirement analysis activity is the process of analysing system requirement based on the list of needs collected in previous activities. The method which is used to asses the suitable candidates for FMADM cases are TOPSIS dan Weighted Product. The basic concept of TOPSIS method is that the best alternative not only has the shortest distance from positive ideal solution, but also has the longest distance with negative ideal solution. Weighted product (WP) is a standard form of FMADM. That concept has been used widely in several MADM model to solve a problem practically [16].

### 2.1. FMADM
In General, the fuzzy multiple attribute decision making procedure follows these steps:

Step 1**:** Set a number of alternatives and some attributes or criteria.
Decision-makers determine some alternatives that will be selected following several attributes or criteria. For example S = {S1, S2, ..., Sm} is the set of alternative; K = {K1, K2, ..., Kn} is the set of attribute or criteria, and A = {aij | i=1,2,...,m; j=1,2,...,n} is the matrix decision where aij is the numerical value of alternative i for attribute j.

Step 2: Evaluation of Fuzzy Set
There are two activities at this step:
a) Choosing a set of rating for the weight of criteria and the degrees of suitability for each alternative with the criteria.
b) Evaluating the weight of criteria and degree of suitability for each alternative with the criteria.

### 2.2. TOPSIS Method
In general, the TOPSIS method procedure follows these steps:

Step 1: The Normalized fuzzy decision matrix
In TOPSIS, the performance of each alternative needs to be graded with equation 1.

$$r_{ij} = \frac{x_{ij}}{\sqrt{\sum_{i=1}^{m} x_{ij}^2}} \text{ ; with x= decision matrix; i=1,2, … ,m; and j=1,2, … ,n.} \quad (1)$$

Step 2: The weighted normalized fuzzy decision matrix
Positive ideal solution $A^+$ and negative ideal solution $A^-$ can be determined based on the weighted normalized rating ($y_{ij}$) as:

$$y_{ij} = w_i r_{ij}, \text{ with i=1,2,…,m; and j=1,2,…,n.} \quad (2)$$

Step 3: Determining positive and negative ideal solution

Positive ideal solution matrix is calculated with equation 3, whereas the negative ideal solution matrix based on equation 4.

$$A^+ = (y_1^+, y_2^+, ..., y_n^+); \qquad (3)$$

$$A^- = (y_1^-, y_2^-, ..., y_n^-); \qquad (4)$$

Step 4: The distance of each candidate from positive and negative ideal solution

The distance between alternative $A_i$ with positive ideal solution can be formulated with equation 5:

$$D_i^+ = \sqrt{\sum_{j=1}^{n}(y_i^+ - y_{ij})^2} \; ; \; i=1,2,...,m. \qquad (5)$$

The distance between alternative $A_i$ with negative ideal solution can be formulated with equation 6:

$$D_i^- = \sqrt{\sum_{j=1}^{n}(y_{ij} - y_i^-)^2} \; ; \; i=1,2,...,m. \qquad (6)$$

Step 5: Determining the value of preference for each alternative
The preference value for each alternative ($V_i$) is given as:

$$V_i = \frac{D_i^-}{D_i^- + D_i^+} \; ; \; i=1,2,...,m. \qquad (7)$$

## 2.3. WP Method

In general, the FMADM weighted product procedure follows these steps:
Step 1: The Normalized fuzzy decision matrix

The WP method uses multiply to relate attribute rating, in which each of it has to be powered with its associated weight.

Step 2: In WP, the performance of each alternative $A_i$ needs to be grading with equation 8.

$$S_i = \prod_{j=1}^{n} x_{ij}^{w_j} \; ; \; \text{with i= 1,2,...,m.} \qquad (8)$$

where $\sum w_j$ = 1. $w_j$ is the power with positive value for advantage attribute, and with negative value for cost attribute.

Step 3: The relative preference for each alternative is given as:

$$V_i = \frac{\prod_{j=1}^{n} x_{ij}^{w_j}}{\prod_{j=1}^{n} (x_j^*)^{w_j}} \; ; \; \text{with i= 1, 2,...,m.} \qquad (9)$$

## 3. Results and Analysis

The scholarships come from government agencies, State Enterprises and several private foundations that concern with the advancement of education. Annually, the Ministry of Religious Affairs offers academic and non academic scholarships for students at UIN.

### 3.1. Requirement Gathering and Modelling
#### 3.1.1 Requirement Gathering

Requirements gathering activities were aimed to analyze the scholarship selection process at the Faculty of Science and Technology. The process was described in the activity diagram shown in Figure 2.

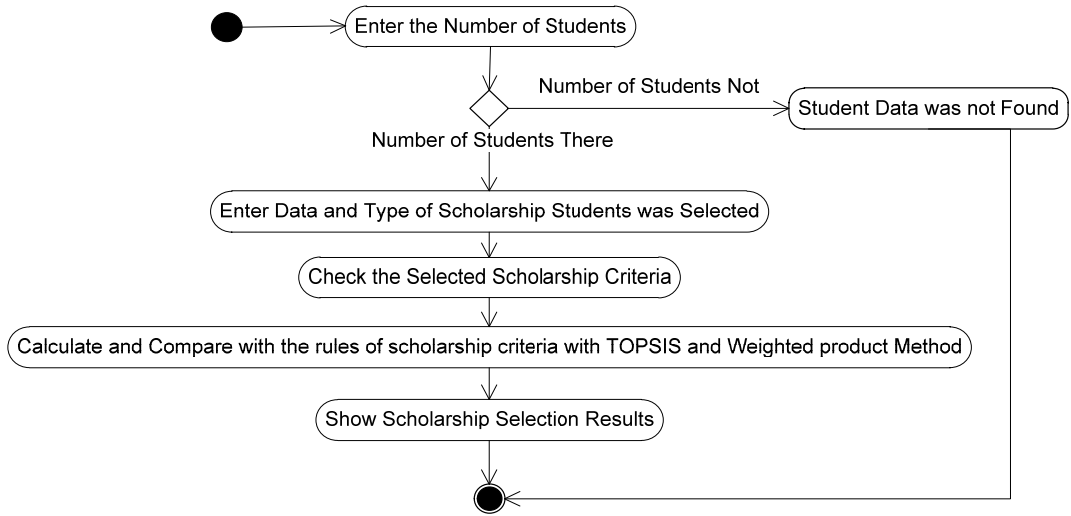

Figure 2. Activity Diagram for Assessing the Feasibility of Scholarship

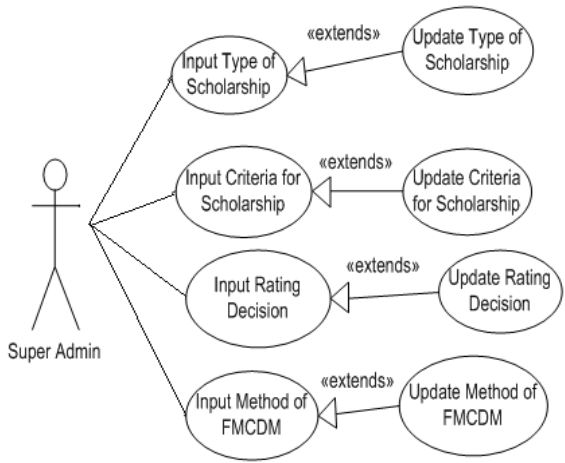

Figure 3. Use case diagram for super admin user

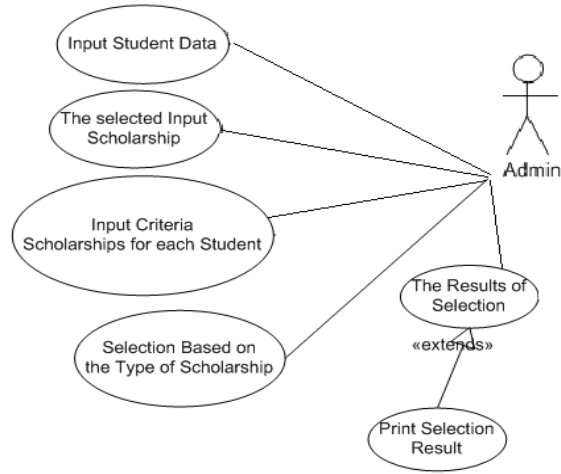

Figure 4. Use case diagram for user admin

#### 3.1.2 Actor and Use Case
The use case diagram for each actor was:
a) Super Admin User
   Super-admin user was a user with an authority to input and up-date data on the system. Super admin input data about the type, criteria and rating decision scholarships that were used for scholarship selection. Use case diagram for a super admin user is shown in Figure 3.

b) Admin User

Admin user is a user whose task is to select students who were applying for a scholarship. Admin can input student data and the type of scholarship and its criteria. The system will display the results of the scholarship selection using TOPSIS method. Use case diagram for an admin user is shown in Figure 4.

**3.1.2 Sequence Diagram**

A sequence diagram in Unified Modeling Language (UML) is a kind of interaction diagram that shows how processes work and in what order. The sequence diagram for scholarship selection is shown in Figure 5.

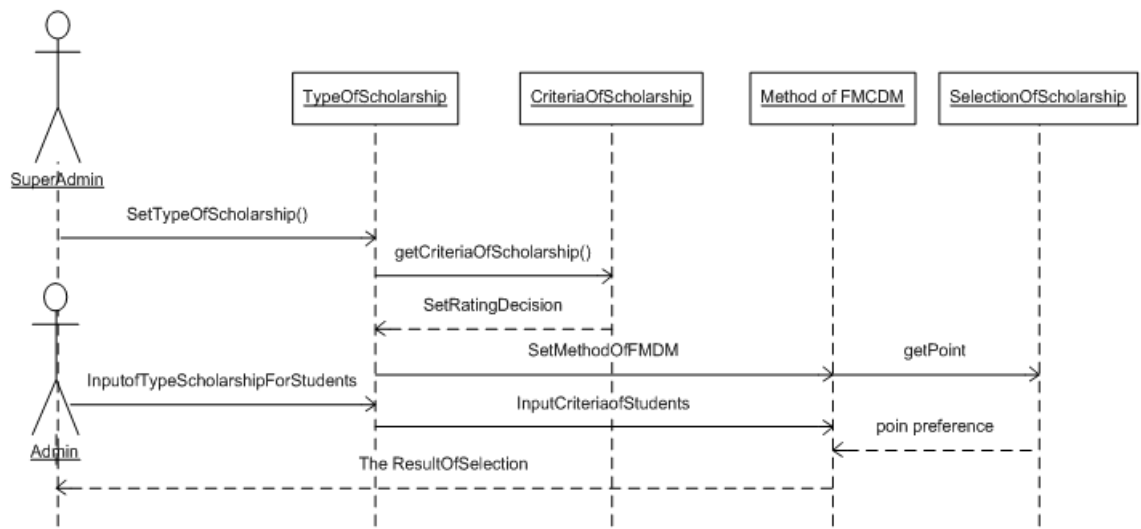

Figure 5. Sequence diagram

**3.2. Requirement Analysis for Fuzzy Multiple Attribute Decision Making**

The proposed method which is applied to solve this problem and the computational procedure were summarized as follows:

Step 1: Set a number of alternatives and some attributes or criteria.

There were 3 criteria used as a basis for decision making in academic scholarship. The criteria include:
- C1 = cumulative grade point;
- C2 = income / economic parents;
- C3 = number of family members

As for the preference, academic scholarship was given to students with a good academic achievement, and coming from a low class with a big family member. On the other hand, there were 9 criteria used to select the candidates for non academic scholarship. These criteria are consisted of:
- C1 = cumulative grade point;
- C2 = income / economic parents;
- C3 = number of family members;
- C4 = religious and moral aspects of Pancasila;
- C5 = aspects of reasoning and idealism;
- C6 = aspects of leadership and loyalty;
- C7 = aspects of interests, talents and skills;
- C8 = aspects of professional activities / internships;
- C9 = aspects of community service;

The preference for non-academic scholarship recipients was students who had creative achievements and joined in extracurricular activities. The administrative requirements for students to get a scholarship were: Indonesian citizen, active student; passed the Sosialisasi Pembelajaran (introductory academic) program at UIN Sunan Kalijaga; passed user education with a value between 60-74; cumulative grade point ≥ 3,0; at least the 3rd semester student; not receiving a scholarship from another sponsor at the moment and enclosed a certificate of good conduct. The purpose of this decision was to find the best three candidates for the scholarship based on specific criteria. There were 15 people (alternate) who passed the administration and surpassed the passing grade given in certain condition: MH01, MH02, MH03, MH04, MH05, MH06, MH07, MH08, MH09, MH10, MH11, MH12, MH13, MH14 dan MH15.

Step 2: Evaluation of Fuzzy Set

The choosing oaf a set of ratings for criteria weights and degrees of suitability of each alternative is based the criteria. Top of form linguistic variables represented the weight of decision for each attribute (criterion). The decision for scholarship criteria was graded as it shown in Table 1. The decision for non academic scholarship criteria was graded as it shown in Table 2.

Table 1. Linguistic variables for the importance weight of each criterion

| Criteria | Linguistic Variable | Fuzzy Number |
|---|---|---|
| C1 | Very High (VH) | (0.75, 1.00, 1.00) |
| C2 | High (H) | (0.50, 0.75, 1.00) |
| C3 | Medium (M) | (0.25, 0.50, 0.75) |

Table 2. linguistic variables for the importance weight of each criterion

| Criteria | Linguistic Variable | Fuzzy Number |
|---|---|---|
| C1 | Medium (M) | (0.25, 0.50, 0.75) |
| C2 | Very Low (VL) | (0.00, 0.25, 0.50) |
| C3 | Very Low (VL) | (0.00, 0.25, 0.50) |
| C4 | Very High (VH) | (0.75, 1.00, 1.00) |
| C5 | Very High (VH) | (0.75, 1.00, 1.00) |
| C6 | Very High (VH) | (0.75, 1.00, 1.00) |
| C7 | High (H) | (0.50, 0.75, 1.00) |
| C8 | Very High (VH) | (0.75, 1.00, 1.00) |
| C9 | High (H) | (0.50, 0.75, 1.00) |

All criteria used fuzzy data except for the first and third criterion. Cumulative grade point and number of family members used crisp data. Criteria of income / economic parents (C2) had compatibility degree with some alternatives decision: T (Compatibility) = {S, F, B}. Membership function for each element was represented using triangular fuzzy numbers with S = small with fuzzy numbers (0.10, 0.10, 0.50); F = Fair with fuzzy numbers (0.00, 0.50, 0.90) and B = Big with fuzzy number (0.50, 0.90, 0.90).

On the other hand, the criteria of C4, C5, C6, C7, C8 and C9 have compatibility degree with some alternatives decision: T (Compatibility) = {VP, P, F, G, VG}. }. Membership function for each element is represented using triangular fuzzy numbers with VP = Very Poor with fuzzy numbers of (0.00, 0.00, 0.25); P = Poor with fuzzy numbers (0.00, 0.25, 0.50) ; F = Fair with fuzzy numbers at (0.25, 0.50, 0.75); G = Good with fuzzy numbers (0.50, 0.75, 1.00) and VG = Very Good with fuzzy numbers (0.75, 1.00, 1.00). Weights for the criteria and degrees of suitability of each alternative were evaluated with the criteria. Decision criteria given by decision makers were graded to assess the eligibility of scholarship recipients. The degree of suitability criteria and decision alternatives were shown in Table 3.

### 3.3. TOPSIS Method

Data on Table 3 were first normalized using equation 1 in order to obtain normalized matrices for both academic and non-academic scholarships. A normalized weight of fuzzy decision was then calculated based on equation 2 following the previous step. Positive ideal solution ($A^+$) is calculated by equation 3. While negative ideal solution ($A^-$) was calculated using equation 4 for each type of scholarship. The result of academis scholarship is shown in Table 4, while the result of non academic scholarship is shown in Table 5.

Table 3. The final aggregated results obtained from grading the numerical example presented in this paper by decision makers

| Alternative | C1 | C2 | C3 | C4 | C5 | C6 | C7 | C8 | C9 |
|---|---|---|---|---|---|---|---|---|---|
| MH1 | 3,22 | S | 7 | VP | VP | VG | G | VG | F |
| MH2 | 3,34 | F | 3 | VP | VP | VG | G | VG | F |
| MH3 | 3,51 | F | 3 | F | VP | VG | G | VG | F |
| MH4 | 3,48 | B | 2 | F | P | P | G | P | VG |
| MH5 | 3,77 | B | 2 | P | P | VP | F | P | VG |
| MH6 | 3,80 | B | 3 | G | F | VP | F | VP | VG |
| MH7 | 3,50 | F | 4 | G | F | F | F | VP | VG |
| MH8 | 3,00 | F | 5 | VG | VG | F | VP | G | B |
| MH9 | 3,12 | S | 4 | F | VG | F | VP | G | B |
| MH10 | 3,90 | S | 3 | VP | G | VG | VP | G | B |
| MH11 | 3,58 | B | 2 | P | G | G | VG | G | B |
| MH12 | 3,72 | B | 2 | G | G | G | VG | VP | VP |
| MH13 | 3,12 | B | 1 | VG | VP | VP | G | VG | P |
| MH14 | 3,01 | S | 3 | F | P | P | G | VG | P |
| MH15 | 3,92 | F | 4 | F | F | VP | VP | VG | F |

Table 4. positive and negative ideal solutions for academic scholarship

|  | $y_1$ | $y_2$ | $y_3$ |
|---|---|---|---|
| Solusi Ideal positif $(y_n^+)$ | 0,291 | 0,363 | 0,258 |
| Solusi Ideal negatif $(y_n^-)$ | 0,223 | 0,040 | 0,037 |

Table 5. positive and negative ideal solutions for non academic scholarship

|  | $y_1$ | $y_2$ | $y_3$ | $y_4$ | $y_5$ | $y_6$ | $y_7$ | $y_8$ | $y_9$ |
|---|---|---|---|---|---|---|---|---|---|
| Solusi Ideal positif $(y_n^+)$ | 0,450 | 0,091 | 0,000 | 0,444 | 0,465 | 0,408 | 0,303 | 0,346 | 0,276 |
| Solusi Ideal negatif $(y_n^-)$ | 0,110 | 0,010 | 0,000 | 0,444 | 0,000 | 0,000 | 0,000 | 0,000 | 0,000 |

Table 6. Result of TOPSIS for academic and non academic scholarship

| Alternative | Academic Scholarship | | | | Non Academic Scholarship | | | |
|---|---|---|---|---|---|---|---|---|
|  | $D_{positive}$ | $D_{negative}$ | $V_i$ | Rank | $D_{positive}$ | $D_{negative}$ | $V_n$ | Rank |
| MH1 | 0.395 | 0.016 | 0.039 | 15 | 0.668 | 0.593 | 0.472 | 10 |
| MH2 | 0.182 | 0.219 | 0.547 | 8 | 0.664 | 0.599 | 0.474 | 9 |
| MH3 | 0.179 | 0.222 | 0.552 | 7 | 0.541 | 0.639 | 0.542 | 6 |
| MH4 | 0.049 | 0.373 | 0.883 | 4 | 0.581 | 0.464 | 0.444 | 13 |
| MH5 | 0.331 | 0.162 | 0.328 | 15 | 0.699 | 0.374 | 0.348 | 15 |
| MH6 | 0.074 | 0.359 | 0.829 | 6 | 0.613 | 0.521 | 0.459 | 12 |
| MH7 | 0.198 | 0.119 | 0.501 | 10 | 0.502 | 0.555 | 0.525 | 7 |
| MH8 | 0.229 | 0.177 | 0.436 | 11 | 0.385 | 0.753 | 0.661 | 1 |
| MH9 | 0.346 | 0.111 | 0.243 | 14 | 0.450 | 0.646 | 0.589 | 3 |
| MH10 | 0.039 | 0.376 | 0.907 | 1 | 0.567 | 0.632 | 0.572 | 5 |
| MH11 | 0.045 | 0.374 | 0.893 | 3 | 0.348 | 0.661 | 0.632 | 2 |
| MH12 | 0.039 | 0.375 | 0.904 | 2 | 0.482 | 0.652 | 0.575 | 4 |
| MH13 | 0.059 | 0.391 | 0.868 | 5 | 0.658 | 0.616 | 0.484 | 8 |
| MH14 | 0.338 | 0.147 | 0.304 | 13 | 0.567 | 0.499 | 0.468 | 11 |
| MH15 | 0.196 | 0.207 | 0.514 | 9 | 0.619 | 0.495 | 0.444 | 14 |

The distance between alternative $A_i$ with their ideal positive and negative solution are computed using equation 5 and 6 once the ideal solutions are obtained. Preferential value for each alternative ($V_i$) is computed using equation 7. A larger value shows that alternative $A_i$ is preferred. The result of TOPSIS computation is shown in Table 6. Student with the highest value for academic scholarship is MH10 while MH8 is that of highest value for non-academic scholarship according to TOPSIS method.

### 3.4. WP Method

A fuzzy set for each criterion is defined. This set is transformed into its fuzzy number to be normalized in order to obtain its normal weight. The preference for alternative $A_i$ ($S_i$) and the relative preference for each alternative ($V_i$) are shown in Table 7. Student with the highest value for academic scholarship is MH10 while MH8 is that of highest value for non-academic scholarship according to weighted product method.

Table 7. Result of Weighted Product for academic and non academic scholarship

| Alternative | Academic Scholarship | | | Non Academic Scholarship | | |
|---|---|---|---|---|---|---|
| | $S_i$ | $V_i$ | Rank | $S_i$ | $V_i$ | Rank |
| MH1 | 0.15 | 0.07 | 4 | 0.00 | 0.00 | 4 |
| MH2 | 0.07 | 0.03 | 7 | 0.00 | 0.00 | 5 |
| MH3 | 0.08 | 0.04 | 6 | 0.00 | 0.00 | 6 |
| MH4 | 0.06 | 0.03 | 8 | 1.19E-12 | 0.02 | 2 |
| MH5 | 0.07 | 0.03 | 9 | 0.00 | 0.00 | 7 |
| MH6 | 0.05 | 0.02 | 14 | 0.00 | 0.00 | 8 |
| MH7 | 0.06 | 0.03 | 10 | 0.00 | 0.00 | 9 |
| MH8 | 0.04 | 0.02 | 15 | 4.99E-11 | 0.97 | 1 |
| MH9 | 0.26 | 0.13 | 3 | 0.00 | 0.00 | 11 |
| MH10 | 0.43 | 0.22 | 1 | 0.00 | 0.00 | 12 |
| MH11 | 0.07 | 0.03 | 11 | 0.00 | 0.00 | 15 |
| MH12 | 0.07 | 0.03 | 12 | 0.00 | 0.00 | 13 |
| MH13 | 0.11 | 0.05 | 5 | 0.00 | 0.00 | 14 |
| MH14 | 0.33 | 0.17 | 2 | 1.73E-13 | 0.01 | 3 |
| MH15 | 0.06 | 0.03 | 13 | 0.00 | 0.00 | 15 |

### 4. Conclusion

It can be concluded from the results and analysis that modeling using UML in FMADM with TOPSIS and weighted product method can be applied for scholarship selection. Some UML elements were incorporated within this study, such as: activity diagram, use case and sequence diagram. TOPSIS and weighted product can be used for fuzzy and/or crisp data FMADM. A selection based on those methods provide similar product for its first result. The preferential values for both of those methods are nevertheless different due to the differences in their matrices normalization process. A student with the highest value is recommended for the scholarship.

**References**
[1] Terano T. Asai K. Sugeno M. Fuzzy Systems Theory and Its Applications. London: Academic Press. 1992.
[2] Klir GJ, Bo Y. Fuzzy Sets and Fuzzy Logic: Theory and Applications. New York: Prentice Hall, Englewood Cliffs. 1995
[3] Chowdhury S. Champagne P. *Multi Criteria Decision Making in Fuzzy Environment.* The Annual General Conference of the Canadian Society for Civil Engineering. Canada. 2006; GC-073-1: GC-073-10.
[4] Chen Y W. Larbani M. Two-person zero-sum game approach for fuzzy multiple attribute decision making problems. *Science Direct Fuzzy Sets and System*. 2006; 157(1): 34-51.
[5] Kusumadewi S. Pencarian Bobot Atribut pada Multiple Attribute Decision Making (MADM) dengan Pendekatan Objektif Menggunakan Algoritma Genetika (Studi Kasus: Rekruitmen Dosen Jurusan Teknik Informatika Universitas Islam Indonesia). *Jurnal Gematika Manajemen Informatika*. 2005; 7(1): 48-56.

[6]  Kusumadewi S. *Pencarian Bobot Atribut pada Multiple Attribute Decision Making (MADM) dengan Pendekatan Subyektif Menggunakan Algortima Genetika (Studi Kasus: Penentuan Lokasi Gedung).* Seminar Nasional Pendidikan Teknik Elektro. Yogyakarta. 2004; 97-105.
[7]  Ramadan MZ. *A Fuzzy Model For R&D Project Selection with Multi Criteria Decision Making.* Proceedings of the 2nd IIEC. Riyadh Kingdom of Saudi Arabia. 2004; 1-11.
[8]  Zhou D. Ma J. Tian Q. Kwok RCW. *Fuzzy Group Decision Support System for Project Assessment*. Proceedings of the 32nd Hawaii International Conference on System Sciences. Hawaii. 1999: 1-5.
[9]  Cheong CW. Jie LH. Meng MC. Lan ALH. Design and Development of Decision Making System Using Fuzzy Analytic Hierarchy Process. *American Journal of Applied Sciences.* 2008; 5(7): 783-787.
[10] Anand MD. Selvaraj T. Kumanan S. Johnny MA. Application of Multicriteria Decision Making for Selection of Robotic System using Fuzzy Analytic Hierarchy Process. *International Journal Management and Decision Making.* 2008; 9(1): 75-98.
[11] Saghafian S. Hejazi SR. *Multi-criteria Group Decision Making Using a Modified Fuzzy TOPSIS Procedure,* Procedings of the International Confrence on Computational Intelligence for Modelling, Control and Automation, and Internatioanl Conference Intelligent Agents, Web Technologies and Internet Commerce (CIMCA-IAWTIC'05). 2005; 57-62.
[12] Jia X. *Comprehensive Evaluation on Green Productions based on TOPSIS Methodology*. International Conference on Information Management, Innovation Management and Industrial Engineering. 2009: 570-572.
[13] Yuan-guang F. *The TOPSIS Method of Multiple Atribute Decision Making Problem with Triangular-fuzzy-avlued Weight.* International Workshop on Modelling, Simulation and Optimization. 2008: 11-14.
[14] Xu ZS. MAGDM Linear-Programming Models with Distinct Uncertain Preference Structures. *IEEE Transactions on Systems, MAN and Cybernetics*. 2008; 38(5): 1356-1370.
[15] Xu Z. Multiple-Attribute Group Decision Making with Different Formats of Preference Information on Attributes. *IEEE Transactions on Systems, MAN and Cybernetics*. 2007; Vol. 37 (No. 6): 1500-1511.
[16] Kusumadewi S. Hartati S. Harjoko A. Wardoyo R, Fuzzy Multi-Attribute Decision Making (Fuzzy MADM)*, *Yogyakarta: Graha Ilmu. 2006: 87-89.